\documentclass[sn-basic,square,comma,numbers]{sn-jnl}
\setcitestyle{numbers}

\makeatletter
\renewcommand\@biblabel[1]{#1.}
\makeatother
\usepackage{array}
\newcolumntype{P}[1]{>{\centering\arraybackslash}p{#1}}

\usepackage[retain-explicit-plus]{siunitx}
\usepackage{makecell}
\usepackage{lineno}
\usepackage{hyperref}
\usepackage{multirow}
\usepackage{booktabs}
\usepackage{xcolor}

\usepackage{subcaption}
\usepackage{graphicx}

\usepackage{listings}
\usepackage{amssymb}
\usepackage{pifont}

\modulolinenumbers[5]

\usepackage{blindtext}
\usepackage{hyperref}
\usepackage{nameref}
\newcommand{\ViTPrecision}{89}
\newcommand{\TestPrecision}{92.37}
\newcommand{\TestPrecisionBetter}{3.3}
\newcommand{\TestFlopsLess}{63.35}
\newcommand{\TestDataset}{Animals-10}
\newcommand{\TrainSpeedFaster}{2}
\newcommand{\PrecisionImageNet}{72.9}

\newcommand{\TestFlopsLessDS}{72.15}
\newcommand{\PrecisionImageNetDS}{77}

\newcounter{mylabelcounter}

\makeatletter
\newcommand{\labelText}[2]{%
\refstepcounter{mylabelcounter}%
\immediate\write\@auxout{%
 \string\newlabel{#2}{{\unexpanded{#1}}{\thepage}{{\unexpanded{#1}}}{mylabelcounter.\number\value{mylabelcounter}}{}}%
}%
}
\makeatother

\begin{document}

\title[{Compress image to patches for Vision Transformer}]{Compress image to patches for Vision Transformer}

\author{Xinfeng Zhao, Yaoru Sun}


\affil{Tongji University}

\abstract{
The Vision Transformer (ViT) has made significant strides in the field of computer vision. However, 
as the depth of the model and the resolution of the input images increase, 
the computational cost associated with training and running ViT models has surged dramatically.
This paper proposes a hybrid model based on CNN and Vision Transformer, named CI2P-ViT. 
The model incorporates a module called CI2P, 
which utilizes the CompressAI encoder to compress images and subsequently generates a sequence of patches through a series of convolutions. 
CI2P can replace the Patch Embedding component in the ViT model, enabling seamless integration into existing ViT models.
Compared to ViT-B/16, CI2P-ViT has the number of patches input to the self-attention layer reduced to a quarter of the original.
This design not only significantly reduces the computational cost of the ViT model 
but also effectively enhances the model's accuracy by introducing the inductive bias properties of CNN.
The ViT model's precision is markedly enhanced.
When trained from the ground up on the {\TestDataset} dataset, CI2P-ViT achieved an accuracy rate of \TestPrecision\%, 
representing a \TestPrecisionBetter\% improvement over the ViT-B/16 baseline. 
Additionally, the model's computational operations, measured in floating-point operations per second (FLOPs), 
were diminished by \TestFlopsLess\%, and it exhibited a \TrainSpeedFaster\-fold increase in training velocity on identical hardware configurations.
}


\maketitle

\section{Introduction}\label{sec1}

\begin{figure}
    \centerline{\includegraphics[width=\columnwidth]{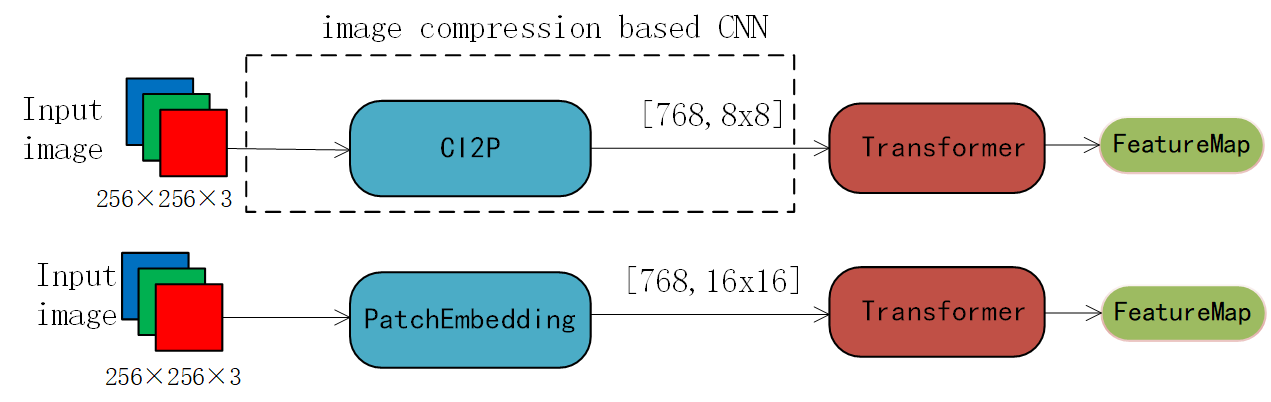}}
    \caption{The CI2P module serves a role analogous to the Patch Embedding component in the standard ViT architecture, 
    yet it generates Patches that are a quarter the size of the original, ensuring minimal loss of visual fidelity. 
    Consequently, this innovation leads to a \TestFlopsLess\% decrease in FLOPs (with images of 256x256 resolution), 
    for the ViT-B/16 model. 
    }
    \label{fig:method_overview}
\end{figure}

In the realm of computer vision, CNNs as referenced in \cite{krizhevsky2017imagenet,lecun1989backpropagation,faster2015towards,long2015fully}, 
have adeptly captured local image features through localized receptive fields coupled with weight sharing, 
showcasing remarkable performance across various computer vision tasks.However, due to the limited receptive field characteristics of CNNs, 
they have limitations in capturing global feature information in images and establishing long-range dependencies.

The Transformer architecture, which has been a game-changer in NLP as documented in \cite{vaswani2017attention,devlin2018bert}, 
features a pivotal mechanism known as self-attention that excels at capturing global dependencies within sequence data. 
The ViT, introduced in \cite{dosovitskiy2020image}, has successfully integrated this self-attention mechanism into image processing, 
unveiling distinctive strengths and considerable potential. 
Specifically, in tasks necessitating a comprehensive understanding of image structure and intricate relational dynamics, ViT has demonstrated remarkable proficiency. 
Nonetheless, ViT is at a disadvantage compared to CNNs due to the absence of inductive biases such as local correlation and translational invariance, 
leading to increased data requirements during training. Moreover, the lack of a down-sampling mechanism in ViT results in higher computational costs, 
as measured by FLOPs.
In response to the computational inefficiency and substantial resource demands of ViT in image tasks, 
researchers have crafted several variants of ViT designed to amalgamate the strengths of both ViT and CNN, thereby enhancing the model's performance and efficiency.

The ViT partitions images into a grid of 16x16-pixel patches, applying the self-attention mechanism to process the image data.
The operational complexity of the self-attention layer (MSA) is directly proportional to both the image dimensions and the model's capacity, 
as depicted by the formula \cite{liu2021swin}:
\begin{align}
\Omega (\text{MSA}) = 4hwC^2 + 2 (hw)^2C
\end{align}
Here, 
C denotes the dimensionality of each image patch, and 
hw is the resultant size after the division into patches. To mitigate the computational demands of ViT, adjustments can be made to either the model's capacity 
C or the spatial extent of the image patches hw.
Reducing C could impair the model's ability to capture complex features, thus the typical strategy involves downscaling the image size hw.
The Swin Transformer \cite{liu2021swin} exemplifies this optimization, employing a hierarchical structure and shifted window mechanism to refine performance by 
incrementally adjusting hw and C at each layer.
CNN-Transformer hybrid models leverage CNN's down-sampling techniques to decrease the spatial dimensions hw while amplifying the channel count, 
thereby reducing the computational overhead of the self-attention mechanism.
For instance, using feature maps derived from a CNN-based Backbone, such as ResNet \cite{he2016deep}, 
and flattening them for input into ViT can notably curtail the computational expense.
However, this approach presents a trade-off: while it minimizes the dimensionality of the image data, it may also result in the loss of essential visual information.
CNN-Backbones are commonly pre-trained on the ImageNet classification task to foster a profound understanding of image content, yielding feature maps enriched with semantic insights.
Yet, these feature maps may omit critical visual details from the original images, 
a drawback that becomes evident when attempting to reconstruct the original image from these feature maps.
For the Vision Transformer, this omitted information could be pivotal, as it may capture distinct semantic features during the processing of the original images.
In pixel-intensive tasks such as small object detection, semantic segmentation, and pose estimation, 
where the preservation of fine-grained details is crucial, this loss of visual information is particularly detrimental.

This research introduces a groundbreaking hybrid model that combines the strengths of CNNs and Transformers, termed CI2P-ViT. 
The CI2P module employs image compression techniques derived from CNNs to diminish the dimensionality of the input to ViT, 
effectively cutting the computational FLOPs required for the self-attention layer by a quarter. 
This reduction is achieved without compromising the model's ability to capture the full spectrum of visual details present in the image, 
as illustrated in Figure~\ref{fig:method_overview}.

The dimension reduction component of the CI2P module is trained in isolation from specific visual tasks, using an encoder-decoder framework. 
This approach ensures that the image is first compressed into a lower-dimensional representation by the encoder and then accurately reconstructed by the decoder, 
preserving as much of the original visual information as possible.

Moreover, the CI2P module integrates CNN's inductive biases into the ViT framework, 
thereby bolstering the model's capacity to detect and process local features within the image. 
A key advantage of CI2P-ViT is that it preserves the original ViT architecture, making it more adaptable for multimodal research and expansion compared to other models, 
such as Swin Transformer, which necessitate alterations to the internal structure of ViT.

This design philosophy not only streamlines the computational efficiency of the model but also broadens its applicability, 
paving the way for future innovations in multimodal analysis and beyond.

The principal contributions presented in this manuscript encompass the following:
\begin{itemize}
\item We propose a CNN-based CI2P module, 
which can be used as a plug-and-play component for the Patch Embedding phase of ViT. 
This method maintains the structure of the ViT model while incorporating the inductive biases of CNNs.
\item We introduce a pioneering fusion of the CNN-based image compression technique, CompressAI, with the ViT framework.
This synergy has led to a reduction in image data dimensionality, which significantly decreased the model's FLOPs.
\item We propose innovative hybrid model named CI2P-ViT.Based on this, a variant with a dual-scale attention mechanism, {CI2P-ViT}$^{ds}$, was designed.
Our model's efficacy is substantiated by experimental outcomes on image classification datasets.
\end{itemize}

Our model enhances both performance and precision without necessitating modifications to the underlying ViT structure. 
When trained from scratch on the {\TestDataset}, our model surpasses the baseline ViT model, achieving an accuracy rate of \TestPrecision\%, 
which corresponds to a \TestPrecisionBetter\% enhancement.
The model also exhibits a \TestFlopsLess\% reduction in FLOPs, leading to a marked reduction in the memory footprint required for ViT model execution. 
This advancement is particularly beneficial for researchers who have constrained resources, 
as it enables more extensive optimization and broadens the potential for further development of the ViT model.

\section{Related Works}\label{sec2}

Within the domain of deep learning, CNNs as introduced in \cite{krizhevsky2017imagenet}, 
have been established as the foundational architecture for a variety of computer vision tasks. 
CNNs adeptly capture localized image features via their convolutional layers, 
securing outstanding achievements in image classification \cite{lecun1989backpropagation}, 
object detection \cite{faster2015towards}, and semantic segmentation \cite{long2015fully}. 
Despite these successes, CNNs encounter challenges in capturing long-range dependencies and integrating global contextual information within images.

The ViT detailed in \cite{dosovitskiy2020image}, represents a novel architectural shift by adopting the Transformer model's \cite{vaswani2017attention} mechanisms 
from NLP and applying them to the computer vision field. 
ViT processes images by segmenting them into patches, treating these as sequence inputs for the Transformer, 
and has demonstrated notable performance across a spectrum of visual tasks. 
This review will concisely examine the models related to CNNs and Vision Transformers, highlighting their evolution, applications, 
and interplay within the landscape of computer vision research.

\subsection{Vision Transformer}
\label{subsec:vit}

The Transformer model, as introduced in \cite{vaswani2017attention}, has achieved remarkable success in NLP.
This success has inspired researchers to explore the application of Transformer concepts in visual tasks. 
Dosovitskiy et al. \cite{dosovitskiy2020image} proposed the Vision Transformer (ViT), a model designed for image recognition.
ViT operates by partitioning an image into a series of patches, which are then processed as sequence data by the Transformer architecture, leading to superior performance in image classification tasks.
Despite these advancements, ViT's training demands a larger dataset and more substantial computational resources compared to CNNs.
The DeiT (Data-efficient Image Transformer) \cite{touvron2021training} addresses this challenge by integrating attention training and knowledge distillation techniques, 
demonstrating that effective training of ViT can be achieved using solely the ImageNet dataset.
DeiT leverages a pre-trained CNN, such as ResNet, in a teacher-student configuration, where the CNN provides guidance to enhance the Transformer's self-attention mechanisms, 
thereby boosting the overall model performance.
The Swin Transformer, an evolution of the ViT, introduces a novel hierarchical structure and a shifted window mechanism, setting it apart from its predecessor \cite{liu2021swin}.
This model optimizes computational efficiency by implementing a local window self-attention mechanism, which confines self-attention calculations to local image windows, 
thereby significantly curtailing the computational demands.
The Swin Transformer has made a mark in the field of computer vision, delivering exceptional results in diverse tasks such as object detection, 
semantic segmentation, image generation, and video action recognition, showcasing its versatility and robustness in handling complex visual challenges.

\subsection{CNN-Transformer Hybrid Model}
\label{subsec:cnn_transformer_hybrid}

The CNN-Transformer hybrid model is an innovative neural network architecture that merges the prowess of CNNs in local feature extraction with the comprehensive 
global context understanding of Transformer models. These models aim to harness the advantages of both architectures to significantly boost performance in visual tasks.

In the Context Vision Transformer (CvT) \cite{hassani2021escaping}, the authors present a distinctive structural approach by incorporating a series of convolutional layers prior 
to each attention mechanism. This design effectively decreases the spatial resolution of feature maps across layers while simultaneously expanding their feature dimensions. 
The authors report that CvT achieves superior performance on the ImageNet-1k dataset with a lower parameter count and reduced computational demand compared to 
the ViT and the DeiT.

Conversely, the Cross-Encoded Image Transformer (CeiT) \cite{yuan2021incorporating} employs an alternative strategy, 
utilizing convolutional layers for initial image downsampling before processing through ViT. 
This method capitalizes on CNNs' proficiency in low-level feature extraction and curtail computational expenses by diminishing the length of the 
image patch sequence subjected to self-attention mechanisms.

\subsection{End to End Image Comparison}
\label{subsec:end_to_end_comparison}

CompressAI, introduced in \cite{begaint2020compressai}, is a PyTorch-based library and evaluation framework 
meticulously crafted to facilitate end-to-end image compression research.
The platform incorporates multiple state-of-the-art end-to-end image compression models\cite{balle2018variational,minnen2018joint,cheng2020learned} that leverage CNNs.
These models have demonstrated compression capabilities competitive with established JPEG and PNG algorithms, 
thereby highlighting the promising prospects of deep learning in advancing image compression technology.
\subsection{Comparison}
\label{subsec:comparison}

ViT and DeiT process images in their original dimensions as inputs to the self-attention layer, 
which results in a significant escalation of computational requirements with increased model complexity and higher image resolutions.
While the Swin Transformer has effectively minimized the number of FLOPs through its pioneering local window mechanism, 
its modifications to the internal structure of ViT might hinder its adaptability for multimodal research and scalability.
Alternative models, such as the CvT, CeiT, and other CNN-Embedding based architectures, 
have adopted the technique of leveraging feature maps derived from CNNs as inputs to the self-attention layer, thereby curtailing computational costs.
Nevertheless, the downsampling inherent in this approach can result in the loss of fine-grained visual information, which is pivotal for tasks that demand high image fidelity, 
such as semantic segmentation and pose estimation. In these pixel-intensive applications, the subtleties of the image are especially critical.

Various approaches within the CNN-Embedding paradigm commonly incorporate feature maps extracted from pre-trained CNN architectures like ResNet-50, 
which are often originally trained on the ImageNet dataset. The efficacy of these models when generalized to datasets outside of ImageNet may be inconsistent, 
thus rendering the performance of ViT model substantially dependent on the proficiency of the CNN-Embedding training process.
Additionally, the CNN-Embedding segment typically necessitates fine-tuning during the ViT training phase. 
This process escalates the computational expense and complicates the attribution of optimization advancements, 
obscuring whether improvements stem from the CNN-Embedding segment or the ViT architecture itself.
The pursuit of novel structure for ViT and the execution of ablation studies are further complicated by this variability. 

In this research, we propose the CI2P-ViT model, integrating end-to-end compression technology derived from CNNs to perform downsampling on input images, 
effectively reducing their dimensions while striving to retain essential visual information. 
This innovative approach allows the ViT to gain a more nuanced understanding of the intricacies present within images.

Distinct from conventional CNN-Embedding techniques, which are often tailored to specific visual tasks, 
the image compression component of CI2P is trained autonomously, enhancing its versatility. 
A notable aspect of the CI2P-ViT model's training regimen is the freezing of the image compression model's parameters. 
This strategic decision minimizes the computational expenses associated with training.

we implement image compression techniques to reduce the volume of image patches processed by the ViT self-attention layer by a factor of four, 
leading to a marked reduction in the model's computational load, measured in FLOPs. 
The integration of the CI2P module introduces CNN inductive biases into the model, enhancing its precision and accelerating the training phase.
When juxtaposed with alternative models like the Swin Transformer and other CVT variants, 
the CI2P-ViT model distinguishes itself by preserving the pristine internal architecture of ViT, eschewing any modifications. 
This strategic approach facilitates the broader application and extension of the Transformer framework within the multimodal research domain.

\section{Method}\label{sec3}

\begin{figure}[t]
    \centering
      \includegraphics[width=\linewidth]{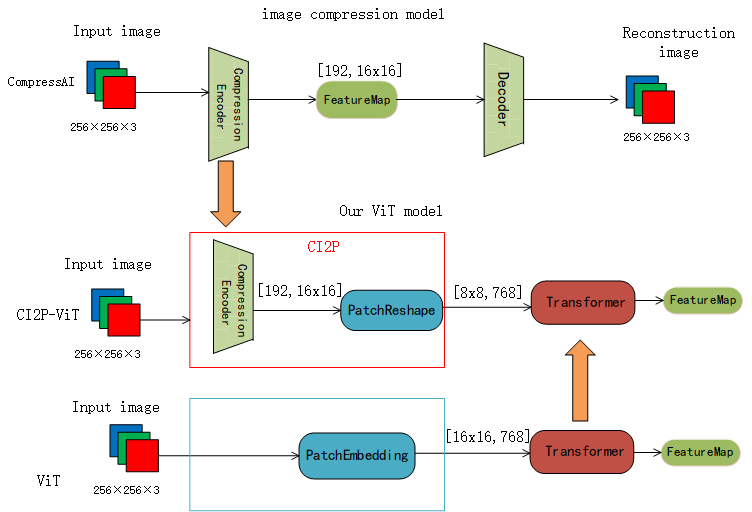}

      \caption{CI2P acts as the Patch Embedding module of the ViT. The Encoder part is the component from the image compression model.}
     \label{fig:ci2p_vit}
  \end{figure}

We present an innovative model termed CI2P-ViT, with its structural layout elegantly illustrated in Figure~\ref{fig:ci2p_vit}. 
The CI2P module is comprised of two key components: the CI2P-Encoder and the CI2P-PatchReshape.
The CI2P-Encoder functions as an image compression unit, integrating a lossy compression model derived from CompressAI\cite{begaint2020compressai}. 
It harnesses the capabilities of a CNN to produce a condensed latent representation that closely mirrors the visual quality of the original image. 
Notably, the training of this component is decoupled from the training of the ViT model, 
rendering the parameters of the CI2P-Encoder frozen throughout the ViT training regimen.
This design strategy guarantees that the improvements observed in the model's capabilities are fundamentally 
due to the intrinsic optimization of the ViT.
The CI2P-PatchReshape module serves as a dimensionality adjustment component, 
tasked with aligning the compressed image output to the input dimensions expected by the ViT model. 

  \subsection{Compress Image to patches}
  \label{sec:method-CI2P}
  The CI2P-Encoder component within the CI2P framework is tasked with the compression of images, employing the CompressAI library, 
  which offers a suite of CNN based end-to-end image compression algorithms. 
  In the CompressAI framework, the loss function is defined as follows:
  \begin{align}
      &y_{[\frac{c*d*d}{s},\frac{h}{d},\frac{w}{d}]} = encoder(x_{[c,h,w]}) \label{ci2p_encoder}\\
      &\hat{y}_{[\frac{c*d*d}{s},\frac{h}{d},\frac{w}{d}]}, y^{*} = quantize(y_{[\frac{c*d*d}{s},\frac{h}{d},\frac{w}{d}]}) \\
      &\hat{x}_{[c,h,w]} = decoder(\hat{y}_{[\frac{c*d*d}{s},\frac{h}{d},\frac{w}{d}]}) \\
      &\mathcal{D}_{MSE} =  (x_{[c,h,w]}- \hat{x}_{[c,h,w]})^2 \\
      &\mathcal{R}_{bpp} =  entropy(y^{*}) \\
      &\mathcal{L} = \lambda \mathcal{D}_{MSE} + \mathcal{R}_{bpp}  \label{CompressAI_loss}
  \end{align}
  Among the available models in CompressAI, 
  CI2P opts for the bmshj2018\_factorized(quality=5) model, as referenced in \cite{minnen2018joint}, utilizing solely its encoder component. 
  This strategic selection is made to achieve an optimal balance between the efficiency of image compression and the preservation of visual fidelity. 
  
  The loss function $\mathcal{L}$\eqref{CompressAI_loss} is composed of two components: 
  $\mathcal{D}_{MSE}$, which calculates the mean squared error between the original and reconstructed images, and 
  $\mathcal{R}_{bpp}$, which evaluates the efficiency of the encoding bitrate.
  The encoder, as referenced by equation \eqref{ci2p_encoder}, produces a downsampled representation 
  $y_{[\frac{c*d*d}{s},\frac{h}{d},\frac{w}{d}]}$ of the original image $x_{[c,h,w]}$, 
  where $c$ is the number of color channels, $h$ and $w$ are the height and width of the image, respectively, 
  and $d$ is the downsampling dimension set to 32. The reduction factor $s$, which is 4 in this paper, dictates how much the image dimensions are decreased. 
  The decoder is capable of reversing this process, reconstructing the original image from the compressed representation 
  $y_{[\frac{c*d*d}{s},\frac{h}{d},\frac{w}{d}]}$ and thus ensuring that all visual information from the original is preserved.
  
  The CI2P-PatchReshape component of the CI2P module handles the necessary dimension reshaping to align the compressed image data with the input requirements of the ViT model.
  \begin{align}
      & x_{[N, D]} = Flatten(PatchReshape(y_{[\frac{c*d*d}{s},\frac{h}{d},\frac{w}{d}]}))  \label{ci2p_reshape}
  \end{align}
  
  The encoder of the bmshj2018\_factorized model produces an output tensor 
  $y_{[\frac{c*d*d}{s},\frac{h}{d},\frac{w}{d}]}$ with dimensions [192, 16x16], which does not align with the input dimensions required by the ViT-B/16 model, 
  specifically 768. To address this mismatch, the CI2P-PatchReshape component is employed to adjust the dimensions accordingly. 
  Post-convolution, the spatial dimensions are halved, and the channel dimension is increased fourfold, yielding an output of [768, 8x8]. 
  This reshaped output is then flattened to match the input requirements of the ViT model.
  Among the various models available in the CompressAI library. 
  the bmshj2018\_factorized model was selected for its superior balance between compression performance and computational efficiency, 
  It is anticipated that ongoing research will yield deep learning encoders that offer even greater compression efficiency with reduced visual loss, 
  thereby further enhancing the capabilities of models like CI2P in future applications.
  
  \subsection{CI2P-ViT}
  \label{sec:method-CI2P-ViT}
  
  In contrast to the conventional ViT architecture, the integration of the CI2P module entails a straightforward replacement of the Patch Embedding component. 
  The output from the CI2P module, as denoted by \eqref{ci2p_reshape}, serves as the input for the self-attention mechanism, which is formulated as follows:
  \begin{align}
      & feature\_map= transformer(x_{[N, D]})
  \end{align}
  
  This self-attention mechanism mirrors the configuration employed by ViT-B/16, characterized by an attention layer dimension of 
  $ D = 768 $, comprising 12 attention heads, and an MLP hidden layer with a dimension of 3072.
  For an input image with dimensions of 256x256, the sequence length N that feeds into the self-attention layer is 256 in the original ViT, 
  whereas in the CI2P-ViT model, it is reduced to 64. This reduction effectively minimizes the FLOPs.
  The feature extraction backbone of CI2P-ViT is culminated by a Global Average Pooling (GAP) operation. 
  Subsequently, a linear classifier is applied to predict the output. 
  The initial segment of the CI2P module, namely CI2P-Encoders, has been pre-trained in isolation and remains frozen during the CI2P-ViT training phase. 
  
  \subsection{CI2P-ViT$^{ds}$ with Dual-Scale Attention Mechanism}
  \label{sec:method-CI2P-ViT-DS}
\begin{figure}
      \centerline{\includegraphics[width=\columnwidth]{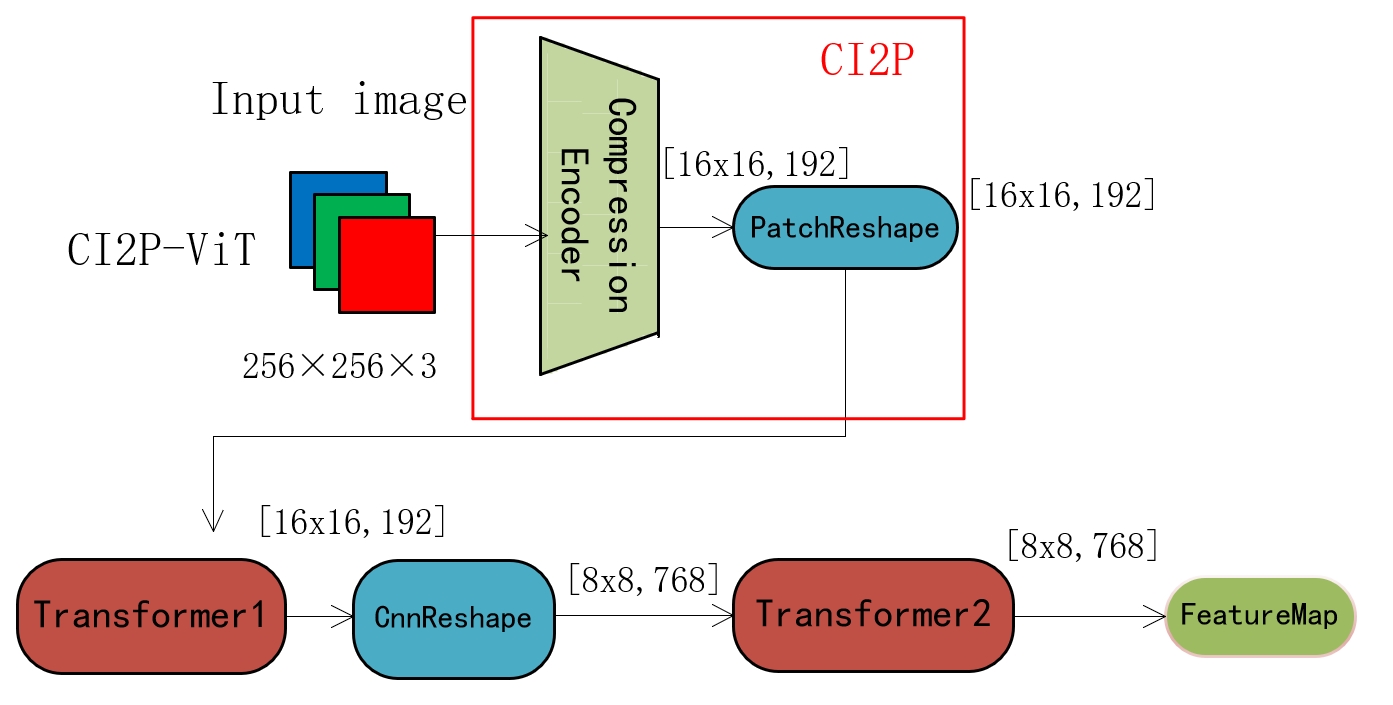}}
      \caption{CI2P-ViT$^{ds}$ with Dual-Scale Attention Mechanism}
      \label{fig:ci2p_vit_ds}
  \end{figure}

\begin{figure}
       \centerline{\includegraphics[width=\columnwidth]{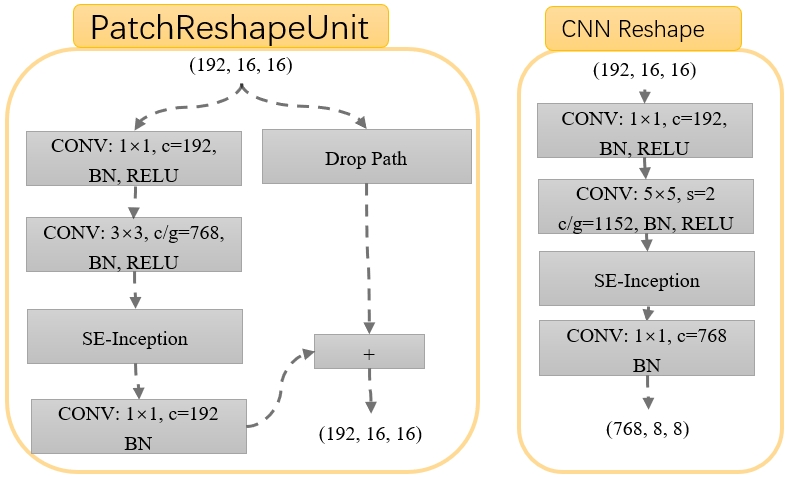}}
       \caption{PatchReshape and CnnReshape.}
       \label{fig:patchreshape}
   \end{figure}

  Compared to the original CI2P-ViT, CI2P-ViT$^{ds}$ incorporates a dual-scale attention mechanism. The output dimension of the CI2P module is set to [192, 16x16], 
  allowing the first six attention layers of ViT to perform attention calculations at the larger 16x16 spatial scale.
  Subsequently, an inverted residual network unit quadruples the dimension while halving the width and height, resulting in a dimension change to [768, 8x8]. 
  The following six attention layers then perform attention calculations at the 8x8 spatial scale.
  The architectural diagram is illustrated in Figure~\ref{fig:ci2p_vit_ds}.
  Both PatchReshape and CnnReshape utilize the inverted residual network units from MobileNet\cite{andrew2017efficient},
  as illustrated in Figure~\ref{fig:patchreshape}.
  
  Compared to CI2P-ViT, the first six attention layers use a dimension of 192, which significantly reduces the model's parameters. 
  The model parameters are reduced from 88.96M to 49.7M, and the FLOPs are decreased from 8.477G to 6.442G. 
  Meanwhile, the accuracy on the ImageNet dataset has increased from {\PrecisionImageNet}\% to {\PrecisionImageNetDS}\%, as detailed in Table~\ref{tab:imagenet_comparison}. 
  Performing attention calculations at two scales, 16x16 and 8x8, allows CI2P-ViT$^{ds}$ to output feature maps at two scales, 
  providing richer semantic information for tasks such as object recognition and pose estimation that heavily utilize multi-scale fusion techniques.

  \section{Experiments}
  \label{sec:experiments}
  
  \subsection{Results on {\TestDataset} Dataset}
  
  \begin{figure}[t]
      \centering
        \includegraphics[width=\linewidth]{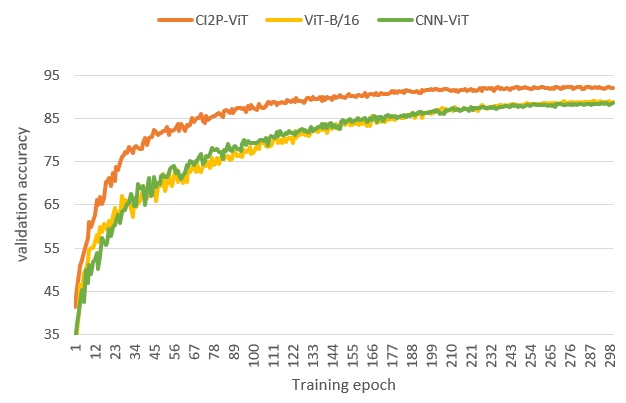}
  
        \caption{{\TestDataset} validation accuracy.}
       \label{fig:ci2p_accuracy3}
    \end{figure}
    
  In our experiments on the {\TestDataset}\cite{animals10}, we trained both the CI2P-ViT and the ViT-B/16 models from scratch. 
  We adopted the mmlab framework. 
  The initial learning rate was set to 1e-04. The parameters for the Adam optimizer were  $\beta1 = 0.9$ and $\beta2 = 0.999$.
  We used random flipping transformation with a probability of 0.5.
  Under identical hardware settings and training parameters, CI2P-ViT completed 300 epochs in approximately 9 hours, whereas ViT-B/16 required about 20 hours, 
  indicating a significant reduction in training duration.
  As depicted in Figure~\ref{fig:ci2p_accuracy3}, 
  CI2P-ViT achieved an accuracy of \TestPrecision\% on the test set, outperforming the ViT-B/16, which attained an accuracy of \ViTPrecision\%.
  
  Unlike conventional CNN+ViT models, the CI2P module strives to preserve as much visual information as possible from the image while performing downsampling.
  To verify the effectiveness of the CI2P design, we constructed a control model CNN-ViT, 
  which uses the same CNN architecture as CI2P but does not use the pretrained parameters from the CompressAI model.
  In CNN-ViT, all parameters of the CNN are trained with the ViT model. After 300 epochs of training, CNN-ViT achieved a maximum accuracy of 88.5\%, 
  which is 3.86\% lower than the accuracy of CI2P-ViT.
  The experimental results indicate that using feature maps from CNN downsampling alone as input for ViT is not sufficient to ensure the improvement of model accuracy.
   The CI2P module, by introducing the inductive bias capabilities of CNNs and preserving visual information of the image, can effectively harness the potential of ViT, 
   thereby achieving better performance in visual recognition tasks.
   
  \subsection{Results on the ImageNet Dataset}

  \begin{table}[t]\centering
      \caption{ImageNet Top-1 validation accuracy
      }
      \footnotesize
          \vspace{2mm}
          \resizebox{\linewidth}{!}{
          \begin{tabular}{l|c|c|c}
              \hline
              \textbf{Model} & \textbf{Params(M)} & \textbf{FLOPs(G)} & \textbf{Top-1(\%)} \\
              \hline
  
              \textbf{DeiT-B \cite{touvron2021training}} & $86$ & $17.6$  & $81.8$  \\
              \textbf{T2T-ViT-24 \cite{2021Tokens}} & $64.1$ & $14.1$  & $82.3$  \\
              \textbf{TNT-B \cite{han2021transformer}} & $65.6$ & $14.1$  & $82.9$  \\
              \textbf{CPVT-B \cite{2021Conditional}} & $88$ & $17.6$  & $82.3$  \\
              \textbf{Swin-B \cite{liu2021swin}} & $88$ & $15.4$  & $83.3$  \\
              \hline
              \textbf{ViT-B/32} & $86$ & $23.13$ & $73.38$   \\
              \textbf{ViT-B/16} & $86$ & $23.13$ & $77.91$   \\
              \textbf{CI2P-ViT} & $88.96$ & $8.477$ & $72.9$   \\
              \textbf{CI2P-ViT$^{ds}$} & $49.7$ & $6.442$ & $77$   \\
              \hline
          \end{tabular}}
          \label{tab:imagenet_comparison}
                \vspace{-5mm}
      \end{table}

  The CI2P-ViT model, initiated without any pre-existing training data and after a comprehensive training regimen of 300 epochs on the ImageNet dataset, 
  realized a test accuracy of \PrecisionImageNet\%. Compared with the performance of current mainstream ViT models, 
  the specific comparative data can be found in Table~\ref{tab:imagenet_comparison}.
  
  While the CI2P-ViT model's accuracy on the ImageNet dataset falls short of some evolved ViT variants,
  its performance closely mirrors the initial ViT-B/32 \cite{dosovitskiy2020image} model's accuracy.
  During the training phase, CI2P-ViT incorporated a limited data augmentation approach, restricted to the application of random flipping.
  Pre-training on expansive datasets such as ImageNet-21K was not part of CI2P-ViT's regimen, nor was training conducted with higher-resolution imagery.
  The CI2P module, tasked with the embedding of image patches, operates with an encoder whose parameters remain static throughout the training process. 
  Although the image encoder's efficacy on ImageNet has not reached its peak, 
  we expect notable enhancements in the CI2P module's capabilities with the ongoing progression of end-to-end image compression technologies.

  \subsection{Performance Comparison}
  
  \begin{table}[t]\centering
      \caption{FLOPs comparison. 
          }
      \footnotesize
          \vspace{2mm}
          \resizebox{\linewidth}{!}{
          \begin{tabular}{l|c|c|c}
              \hline
              \textbf{Image Size} &   \textbf{ViT FLOPs} & \textbf{CI2P-ViT FLOPs} & \textbf{CI2P-ViT$^{ds}$ FLOPs} \\
              \hline
  
              $256^2$ &  $23.127$ G & $8.477(63.35\%\downarrow)$ G  & $6.442(72.15\%\downarrow)$ G\\
              $384^2$ &  $55.433$ G & $19.284(65.21\%\downarrow)$ G & $14.492(73.86\%\downarrow)$ G\\
              $512^2$ &  $107$ G & $34.81(67.47\%\downarrow)$ G & $25.762(75.92\%\downarrow)$ G\\
              \hline
          \end{tabular}}
          \label{tab:FLOPs_comparison}
                \vspace{-5mm}
      \end{table}
  
      The model parameters for ViT-B/16 are 86M, while the model parameters for CI2P-ViT are 88.96M. 
      Due to the addition of the CI2P module, there is a slight increase in model parameters, but there is a significant reduction in FLOPs. 
      Specifically, when the input image size is 256x256 pixels, 
      CI2P-ViT achieves a significant reduction of \TestFlopsLess\% in FLOPs compared to ViT-B/16. 
      As the number of model layers increases and the input image size expands, 
      the effect of reducing FLOPs brought by the CI2P module becomes more pronounced. 
      This performance advantage is detailed in Table~\ref{tab:FLOPs_comparison}. 
      The model parameters for CI2P-ViT$^{ds}$ are 49.7M, which is a 42\% decrease compared to ViT-B/16, 
      and the FLOPs are 6.442G, which is a reduction of \TestFlopsLessDS\%, achieving a very good lightweight effect.

  \section{Conclusion}
  This paper presents the CI2P module, designed to supplant the Patch Embedding segment of the traditional ViT architecture.
  The CI2P module employs CNN-based lossy compression techniques to dimensionally reduce image data while preserving nearly all visual information. 
  This design significantly reduces the computational complexity of ViT's self-attention layer without sacrificing the model's accuracy.
  In fact, the CI2P module slightly enhances the accuracy of ViT by introducing the inductive bias of CNNs.
  Given the expanding dimensions of images in computer vision applications and the escalating complexity of ViT models, 
  the potential for CI2P technology is immense, suggesting its pivotal role in delivering more efficient and precise solutions for visual tasks.

\bibliography{references}
\end{document}